\documentclass[11pt]{article}
\usepackage{amssymb, amscd, amsthm, amsfonts}
\usepackage{graphicx}
\usepackage{hyperref}

\usepackage{authblk}
\usepackage{stackengine}
\usepackage{amsmath}
\usepackage{amsfonts}
\usepackage{tabularx,booktabs}
\newcolumntype{Y}{>{\centering\arraybackslash}X}
\newcolumntype{C}[1]{>{\centering\arraybackslash}p{#1}} 

\def\keywordname{{\bfseries \emph Keywords}}%
\def\keywords#1{\par\addvspace\medskipamount{\rightskip=0pt plus1cm
		\def\and{\ifhmode\unskip\nobreak\fi\ $\cdot$
		}\noindent\keywordname\enspace\ignorespaces#1\par}}

\title{Rank the triplets: A ranking-based multiple instance learning framework for detecting HPV infection in head and neck cancers using routine H\&E images}
\author[1]{Ruoyu Wang}
\author[2]{Syed Ali Khurram}
\author[1]{Amina Asif}
\author[3]{Lawrence Young}
\author[1,4]{Nasir Rajpoot}

\affil[1]{\stackunder{Tissue Image Analytics Centre, Department of Computer Science, University of Warwick}{\tt\small \{ruoyu.wang.2, amina.asif, n.m.rajpoot\}@warwick.ac.uk}}
\affil[2]{\stackunder{School of Clinical Dentistry, University of Sheffield}{\tt\small s.a.khurram@sheffield.ac.uk}}
\affil[3]{\stackunder{Warwick Medical School, University of Warwick}{\tt\small l.s.young@warwick.ac.uk}}
\affil[4]{The Alan Turing Institute, London}

\date{}

\begin{document}
	
\maketitle
	
\begin{abstract}
	The aetiology of head and neck squamous cell carcinoma (HNSCC) involves multiple carcinogens such as alcohol, tobacco and infection with human papillomavirus (HPV). As the HPV infection influences the prognosis, treatment and survival of patients with HNSCC, it is important to determine the HPV status of these tumours. In this paper, we propose a novel triplet-ranking loss function and a multiple instance learning pipeline for HPV status prediction. This achieves a new state-of-the-art performance in HPV detection using only the routine H\&E stained WSIs on two HNSCC cohorts. Furthermore, a comprehensive tumour microenvironment profiling was performed, which characterised the unique patterns between HPV+/- HNSCC from genomic, immunology and cellular perspectives. Positive correlations of the proposed score with different subtypes of T cells (e.g. T cells follicular helper, CD8+ T cells), and negative correlations with macrophages and connective cells (e.g. fibroblast) were identified, which is in line with clinical findings. Unique gene expression profiles were also identified with respect to HPV infection status, and is in line with existing findings.
\end{abstract}


\section{Introduction}
Head and neck squamous cell carcinoma (HNSCC) is the most common type of head and neck cancers (HNC) contributing to more than 550,000 new cases and 380,000 deaths worldwide every year \cite{2019EpidHNSCC}. HNSCC usually locates in the mucosal epithelium of multiple anatomical sites within the head and neck regions \cite{2020JohnsonHNSCC, 2016mayoHNSCC}. The aetiology of this disease has long been thought to be the exposure to carcinogens such as alcohol, tobacco, areca nut or airborne pollutants \cite{2020JohnsonHNSCC}. However, human papillomavirus (HPV) infection has been implicated in the carcinogenic process in a subset of HNSCC cases, which exhibits different epidemiology, disease course, prognosis and morphological profile comparing to the traditional HPV- HNSCC \cite{2012Westra}. The HPV+ HNSCC displays a significantly better prognosis, with the overall survival rate (3-year) of 82\%, comparing to HPV- HNSCC with only 57\% \cite{WHO2017HNC}. Interestingly, with a decrease in tobacco and alcohol consumption globally, the epidemiology of this cancer has, in fact, shifted over the past two decades from conventional HPV- HNSCC to HPV+ HNSCC \cite{2019EpidHNSCC}. 

This shift in epidemiology has posed a challenge in treatment design. 
Because of the unfavourable prognosis of conventional HPV-negative HNSCC, a combination of intensive surgical and chemoradiotherapy treatments had been given to HNSCC patients in the past, regardless of their HPV infection status. These aggressive treatments can cause long-term side effects and reduce the quality of life for some patients \cite{mirghani_treatment_2015}. Many studies \cite{ chen_reduced_dose_2017, mirghani_treatment_2015, 2016mayoHNSCC} have shown the possibility and necessity for patient stratification and treatment de-escalation for HNSCC patients based on HPV presence, which aims at reducing the treatment toxicity while maintaining good cure rates. Consequently, determining patient's HPV infection status becomes a priority in HNSCC diagnosis as it can provide more information on prognosis and treatment design.

\begin{figure}[!htbp]
	\begin{center}
		\includegraphics[width=0.7\textwidth]{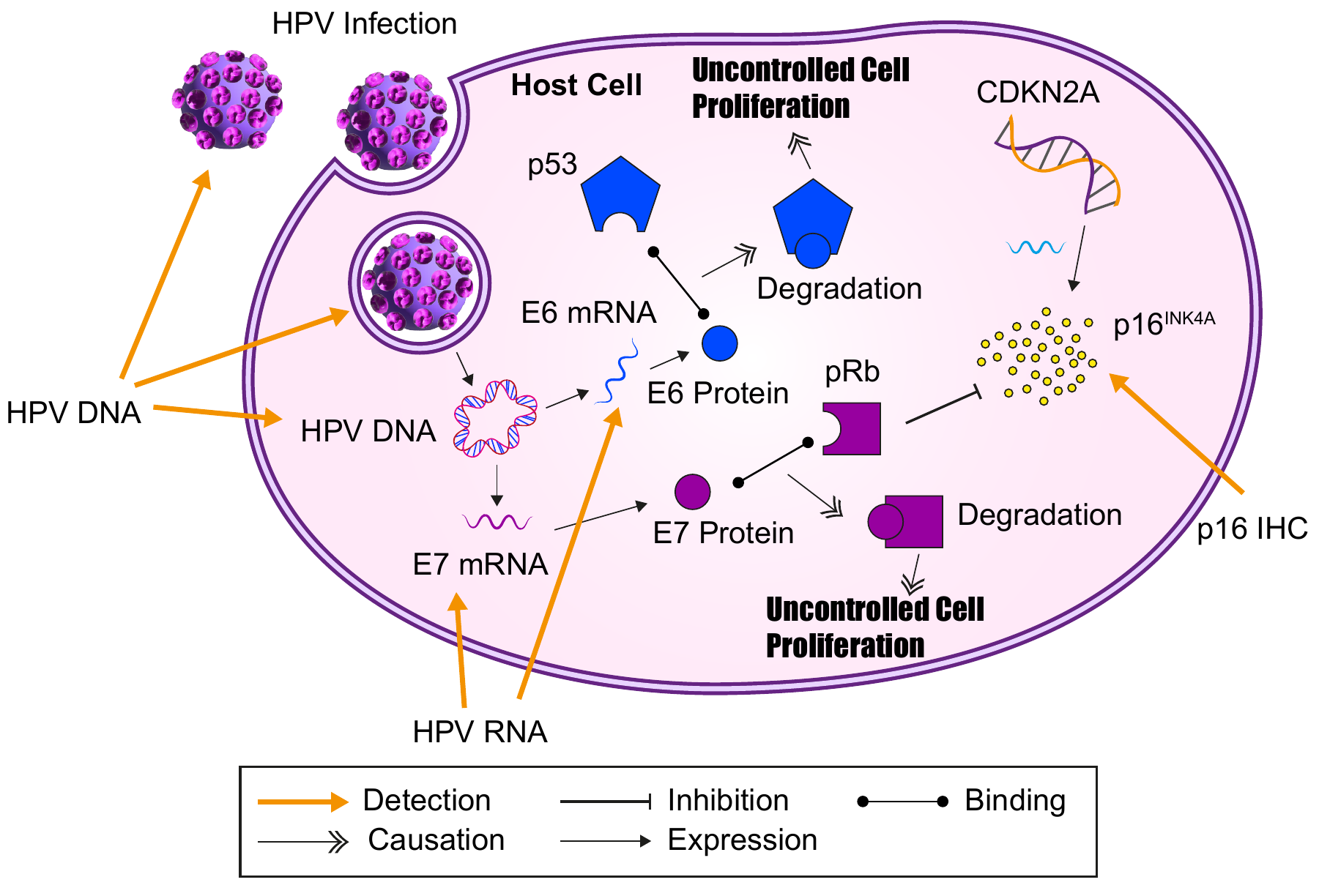}
	\end{center}
	\caption{A simplified illustration of the HPV infection mechanism. It mainly involves two HPV oncoproteins E6 and E7, which can deactivate tumour suppressors p53 and pRb in human cells, leading to an uncontrolled cell proliferation.}
	\label{fig:HPVMechanism}
\end{figure}

Unfortunately, HPV infection status determination is not as easy as it sounds, due to the complex molecular mechanism of the HPV-induced carcinogenesis, which mainly involves two HPV-encoded oncoproteins E6 and E7. As shown in Fig \ref{fig:HPVMechanism}, in HPV-infected cells, E6 proteins facilitates the degradation of the tumour suppressor p53, which leads to malignant cell transformation and uncontrolled cell proliferation. E7 proteins deactivates the tumour suppressor pRb, triggering the desilencing of CDKN2A-encoded protein p16 due to a pRb-mediated negative regulation \cite{molecular_mechanism_HPV}. Therefore, in practice, detecting high-level of p16 expression using immunohistochemistry (IHC) is a common approach used as a surrogate marker to determine the presence of HPV E7 (i.e. \textgreater70\% nuclear and cytoplasmic in oropharyngeal specimens \cite{p16_determination}). However, some studies have identified a subgroup of patients which exhibits p16 positivity but have no HPV DNA presence \cite{rietbergen_molecular_p16pos_2014, lewis_p16_2010}. Therefore, for more specific testing and subtyping, in-situ hybridisation (ISH) and polymerase chain reaction (PCR) are used to specifically identify HPV DNA in tissue \cite{2016mayoHNSCC}. Yet the study by \cite{HPV_POS_MRNA_NEG} reported a significant minority of patients with HPV DNA presence but no expression of HPV E6 nor E7 was detected, suggesting that the HPV may have no involvement in the carcinogenesis of such cases. Therefore, none of these assays are sensitive or specific enough to be used alone due to false positive and false negative tests on occasions, in addition to variability in subjective pathological assessment, hence requiring pathologists to use a combination of assays, repeat these assays or get a consensus opinion with larger specialist cancer centres to ensure an accurate diagnosis \cite{leemans_molecular_2018}. This could incur extra costs and delays in diagnosis. Therefore, it is necessary to explore alternative objective approaches to determine the HPV status of HNSCC patients more effectively and efficiently.

On the other hand, characterising the tumour microenvironment (TME) of these two HNSCC subtypes are of equal importance for developing treatments, finding new biomarkers and understanding the carcinogenesis of HPV-induced HNSCC. For instance, the work by \cite{spanos_immune_2009} demonstrates that the immune response is critical in HPV+ tumour clearance, and radiation therapy alone cannot facilitate this process with an impaired immune system. Findings in \cite{wansom_correlation_CD8_2010} suggest that a high proportion of CD8 cells is linked with improved survival and response to induction chemotherapy. Differences in gene expression patterns have also been identified between HPV+/- HNSCCs, which may provide new insights into the management of patients with different types of tumours \cite{gene_expression}. These studies reveal the possibility of a more refined stratification based not only on the HPV presence, but also on the TME, immunological and genomic profiles of HNSCC patients, which can lead to better treatment designs and more appropriate de-escalation strategies.

In this paper, we introduce a simple and effective Multiple Instance Learning (MIL)-based pipeline for HPV prediction, together with a novel triplet ranking loss function. Our method does not require the tumour region annotation, and achieves the state of the art comparing to previous studies. Our contributions are listed as follows:
\begin{itemize}
	\item [1)] We propose a novel triplet-ranking loss function which achieves the new state-of-the-art performance in detecting HPV presence using H\&E WSIs. 
	\item [2)] In addition to our loss function, we propose a deep ranking MIL network for HPV detection.To the best of our knowledge, this is the first time that a ranking-based MIL approach has been applied to solve this problem.
	\item [3)] With our localised predictions and automatic cellular composition analysis, a tumour-microenvironment profiling on H\&E pathology between HPV+/- patients was performed.
	\item [4)] A comprehensive immunological and genomic analysis of TCGA-HNSC cohort was conducted with predicted ranking score, which revealed and validated the link between HPV involvement and molecular characterisation.
\end{itemize}

\section{Related works}
Many of the existing methods for predicting HPV infection status, or more generally, predicting molecular alterations from histopathology images, can be categorised into hand-crafted feature-driven \cite{fouad_ISH_HPV_2021, koyuncu_multinucleation_2021, FLocK2021} and end-to-end data-driven \cite{KatherHPV, KleinHPV, Kather2019MSI, Coudray2018, Bilal2021, SPOPMutation, Campanella2019, ebv_net}. \cite{fouad_ISH_HPV_2021} and \cite{koyuncu_multinucleation_2021} chose to use the morphological features of the epithelium and to quantify the multi-nucleation respectively for HPV prediction and prognostic analysis. These features were chosen based on the clinical knowledge that we have regarding the HPV impact on the morphological patterns of histopathology slides. These two approaches are easily interpretable as they were developed based on domain-specific knowledge. However, they struggle to generalise on other computational pathology tasks since they were designed for HPV detection specifically. Meanwhile, \cite{FLocK2021} used a "soft" hand-crafted feature selection which based only on the morphological features of nuclei detected, and construct graphs to describe the input image. This method can be extended to other problems since it was not designed using domain-specific knowledge. However, the performance of this method depends heavily on the accuracy of the nuclei segmentation model. Moreover, the selection of the hand-crafted features (i.e. different properties of the morphological patterns) also depends on the dataset used, and might not generalise well to external cohorts \cite{hand_crafted_cons}. 

Therefore, many researchers have explored end-to-end data-driven approaches for such problems. Most of these methods \cite{KatherHPV,KleinHPV, Kather2019MSI, Coudray2018, Bilal2021, SPOPMutation, Campanella2019, ebv_net} were developed based on a common methodology in computational pathology, which is divide-and-conquer. Since modern computers cannot process the multi-gigapixel WSIs in whole, we need to divide the WSI into small patches which can be processed by computers. After "conquered" the small patches, we then aggregate the patch-level information back to either slide-level or patient-level for further analysis. When "conquering" the patches, most of these methods assigned slide-level label to the patches to train a classification network, as we do not generally have the localised ground truth for each patch. This strategy is often being referred to as "weak labelling". When generating patient-level predictions, some of these methods used simple statistical aggregations, such as the average or weighted average of all patches' scores \cite{Coudray2018, SPOPMutation,Bilal2021, ebv_net}, the proportion of positively-classified patches \cite{KatherHPV, Kather2019MSI, Coudray2018}, or the sum of the median and the max score \cite{KleinHPV}. The success of these approaches suggests that the molecular-level alterations can be reflected by the routine H\&E pathology, and the combination of the weak labelling training strategy and the simple statistical aggregations are adequate enough to capture these alterations and make predictions to assist pathologists. However, since we do not have the localised ground truth for each patch regrading the molecular alteration in question, we cannot determine which patch contributes more to the final prediction. Some patches may have more relevance than others, and some patches may show no correlation to the task. Therefore, adopting simple aggregation strategies can lead to suboptimal performance due to the noise it introduced to the overall prediction. Moreover, many of the methods presented here require the analysis to be conducted within the tumour regions of the slides \cite{KatherHPV, KleinHPV, Kather2019MSI, Coudray2018, Bilal2021, SPOPMutation, ebv_net}. This can cause extra manual labours for annotating tumour regions, or to train a tumour detection network which will add uncertainty to the whole pipeline.

Meanwhile, \cite{Campanella2019} explored the use of Multiple Instance Learning (MIL) paradigm for this task. It also adopted a weakly labelling strategy to train a network for patch classification. A recurrent neural network (RNN) was used for aggregating top ranked patches to generate patient-level predictions. Furthermore, CLAM proposed by \cite{CLAM} used an attention network, together with cluster-based constraints for aggregating patch-level features and generating patient-level predictions. These two methods used either RNN or attention network for aggregation, which can learn the relative importance of patches and automatically aggregate the information. However, tumour heterogeneity occurs not only within the same tumour in one patient, but also between the same type of tumours of different patients. The former can cause different patches extracted from the same WSI to contribute differently for the overall prediction, the latter can be reflected by distinct characteristics of the same type of tumours between different patients. The work of \cite{Campanella2019} and \cite{CLAM} addressed the heterogeneity issue in patch-level, but did not take the patient-level heterogeneity into consideration. Both of them used the cross entropy loss to formulate the objective of the model, which ideally would predict a value as close as possible to 1 for positive samples, and 0 for negative samples, indicating their probabilities of being positive. However, this may not be an appropriate way to model the tumour heterogeneity in some computational pathology tasks.

To cope with the aforementioned limitations, we propose the triplet ranking loss and an MIL pipeline for HPV prediction. To take the inter-patient tumour heterogeneity into account, we aims to model the HPV prediction task as a ranking problem, instead of a binary classification problem. Our assumption is that not all patches extracted from the same slide contribute equally to the final prediction, and not all WSIs from different patients exhibit the same degree of HPV-involvement. 

\begin{figure}[!htbp]
	\begin{center}
		\includegraphics[width=0.35\textwidth]{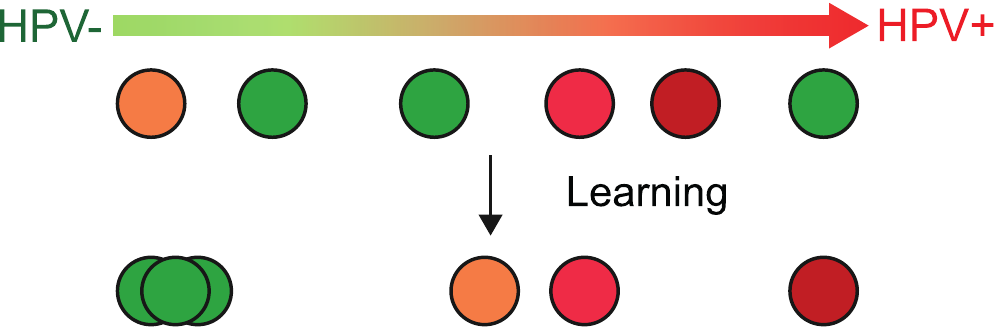}
	\end{center}
	\caption{Intuition of our proposed loss function. Points coloured in green represent HPV- samples, and points coloured in any colour other than green represent HPV+ samples with different degrees of HPV involvement.}
	\label{fig:LossFunction}
\end{figure}

\section{Materials and Methods}
\subsection{Data Collection}
\subsubsection{Whole slide image data}
Data collected from two cohorts were used in this work. 432 diagnostic H\&E-stained whole slide images of 412 patients collected from 26 different medical centres were retrieved from the Head and Neck Squamous Cell Carcinoma program of The Cancer Genome Atlas project (TCGA-HNSC). WSI files were retrieved from the GDC portal(https://portal.gdc.cancer.gov/). HPV infection status for TCGA-HNSC cohort was retrieved from the study by \cite{CAMPBELL2018194}. We adopted the same exclusion criteria as in \cite{KatherHPV}.

\begin{table}[!htbp]
	\begin{center}
		\begin{tabular}{c|c|c}
			\hline
			& TCGA-HNSC & Sheffield \\
			\hline
			No. of Centres & 26 & 1  \\	
			\hline
			No. of HPV$+$ cases & 48 & 40  \\	
			No. of HPV$-$ cases & 364 & 29 \\	
			Total cases & 412 & 69  \\	
			\hline
		\end{tabular}
	\end{center}
	\caption{Details of the dataset used in this work.}
	\label{DataDetails}
\end{table}

For external validation, 72 anonymised diagnostic H\&E-stained whole slide images from 69 OPSCC patients were collected from the University of Sheffield. The HPV status for these cases was determined by routine p16 IHC staining, following the guidance published here \cite{p16_determination}, that is p16 IHC positivity is reported in cases where at least 70\% moderate to strong nuclear and cytoplasmic staining is shown.

Details of these two cohorts are presented in Table \ref{DataDetails}.

\subsubsection{Genomic and immunology data of TCGA-HNSC}
The gene expression data for TCGA-HNSC cohort was retrieved from the GDC portal. The upper quartile normalized RSEM data for batch-corrected mRNA gene expression were used for analysis. The relative fraction of different types of immune cells were estimated using CIBERSORT \cite{cibersort} reported in the study of \cite{THORSSON2018812}.

\begin{figure*}[!ht]
	\begin{center}
		\includegraphics[width=1\textwidth]{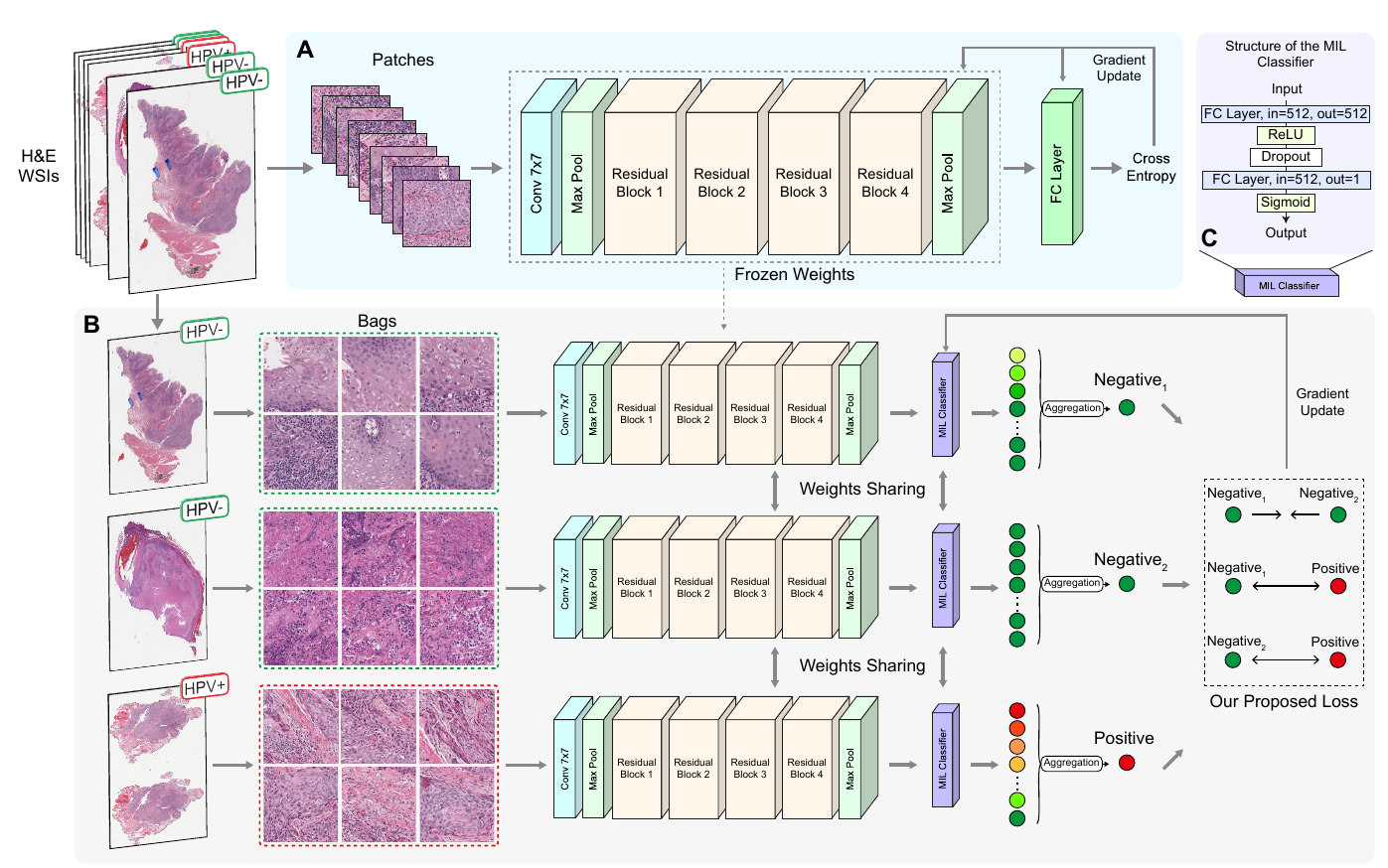}
	\end{center}
	\caption{Our proposed pipeline. \textbf{(A)} Training a feature extractor with a weak labelling strategy. Patches were extracted from WSIs with slide-level labels, and a ResNet-18 was used for training a patch classification model. \textbf{(B)} Our proposed triplet-ranking training pipeline. Patches extracted from the same WSI were grouped into one bag with the slide label. An MIL-classifier was used for scoring each patch. Slide-level score was aggregated from patch-level scores, and the proposed triplet-ranking loss was used for the loss calculation. \textbf{(C)} The detailed structure of the MIL-classifier used.}
	\label{fig:Pipeline}
\end{figure*}

\subsection{Ranking-based Loss}
Unlike Cross Entropy loss or Mean Squared Error loss which aims at learning an accurate mapping between the input data and the ground truth values, ranking-based losses focus more on obtaining a correct order between samples and learning an accurate similarities among given samples. 

Our proposed loss function aims to model the HPV involvement in HNSCC in a weakly supervised fashion. We took our inspiration from the designs of the pairwise ranking loss \cite{ESMIL}, the triplet loss \cite{FaceNet} and the quadruplet loss \cite{Quadruplet}. The pairwise ranking loss has one constraint over two input samples, which is formulated as follows:
\begin{align}
	\label{eqn:ESMIL}
	\begin{split}
		L_{pairwise} & =  \sum_{p,n}^{N} [\alpha - (x_{p} - x_{n})]_+ 
	\end{split}
\end{align}
where $[x]_+ = max(0, x)$, $x_p$ and $x_n$ represent a positive and a negative sample respectively, and $\alpha$ is the margin value. It forces the model to yield higher scores for positive samples over negative ones, with an extra margin parameter to ensure the positive score is higher than the negative score by at least $\alpha$. 

The triplet loss in \cite{FaceNet} imposes two constraints over three input samples. It is formulated as follows,
\begin{align}
	\label{eqn:TripletLoss}
	\begin{split}
		L_{tri} & =  \sum_{i}^{N} \Big[ || f(x_i^a) - f(x_i^p) ||_2^2 - || f(x_i^a) - f(x_i^n) ||_2^2 + \alpha \Big]_+ 
	\end{split}
\end{align}
It forces the model to minimise the distance between the anchor and positive samples, while maximising the distance between the anchor and negative samples. 

\cite{Quadruplet} proposed the quadruplet loss which extended the triplet loss to a case of four input samples. While keeping the same constraints that the triplet loss has, an additional constraint over another negative sample is included in the loss formulation:
\begin{align}
	\label{eqn:QuadrupletLoss}
	\begin{split}
		L_{quad} & =  \sum_{i,j,k}^{N} [g(x_i, x_j)^2 - g(x_i, x_k)^2 + \alpha_1]_+ \\
		& + \sum_{i,j,k,l}^{N} [g(x_i, x_j)^2 - g(x_l, x_k)^2 + \alpha_2]_+ \\
		& (s_i = s_j, s_l \neq s_k, s_i \neq s_l, s_i \neq s_k)
	\end{split}
\end{align}
where $\alpha_1$ and $\alpha_2$ are corresponding margins for each constraint, $g(x_i, x_j)$ is a learned metric which measures the distance between input $x_i$ and $x_j$, and $s_i$ represents the class of image $x_i$. The quadruplet loss takes four inputs. Samples $x_i$ and $x_j$ come from the same class. $x_l$ and $x_k$ comes from two different classes, both of which are different from that of $x_i$. The authors argue that the second term of the equation \eqref{eqn:QuadrupletLoss} can minimise the intra-class variation which leads to a better generalisation.


\subsection{Our proposed loss function}

Our proposed loss function aims to teach the model to quantitatively measure the HPV involvement in HNSCC from mining useful information of H\&E slides.Our assumption is that all the HPV- H\&E WSIs should not exhibit any HPV-induced feature which should only appear in HPV+ H\&E WSIs. Meanwhile, HPV+ H\&E WSIs can present different degrees of HPV-involvement. Therefore, like what is illustrated in Fig. \ref{fig:LossFunction}, we want the scores predicted for HPV+ samples to be higher than that of HPV- samples. However, we do not need all the HPV+ scores to be as high as possible nor close to each other, which is the objective for cross entropy loss or MSE loss. For HPV- scores, we want them to be closer to each other to minimise the intra-class variation for a better generalisation. For HPV+ scores, we do not want to force them to stay close to each other as ideally, the model should be able to represent the heterogeneity among positive samples. However, for better generalisation, we want to maximise inter-class variation between HPV+ and HPV- samples.

Therefore, our proposed loss function is formulated as follows,
\begin{align}
	\label{eqn:OurLoss}
	\begin{split}
		L_{ours} & =  \sum_{i}^{N} [\alpha_1 - (x_i^p - x_i^{n1})]_+  + \sum_{i}^{N} [\alpha_1 - (x_i^p - x_i^{n2})]_+ \\
		& + \sum_{i}^{N} [||x_i^{n1} - x_i^{n2}||_2^2 - \alpha_2]_+
	\end{split}
\end{align}
where  $x_i^p$ denotes the score for HPV+ sample. $x_i^{n1}$ and $x_i^{n2}$ denotes that of any two different HPV- samples. $\alpha_1$ and $\alpha_2$ are the margin parameters for constraining intra-class variation and inter-class variation respectively. Our loss function imposes two types of constraints over three input samples. The first two terms of the equation \eqref{eqn:OurLoss} consider the differences of the predicted values between the HPV+ and two HPV- samples. It forces the model to predict a higher value for HPV+ samples than any of the HPV- samples by at least $\alpha_1$. The third term of equation \eqref{eqn:OurLoss} aims to reduce the distances among HPV- samples to reduce the intra-class variation. 

There are two main differences between our proposed loss function and the three aforementioned ranking-based losses. Firstly, the pairwise ranking loss and the triplet loss do not consider the intra-class variation, which can lead to suboptimal generalisation. Secondly, our loss function considers the absolute position of each predicted score on the coordinate, as illustrated in Figure \ref{fig:LossFunction}. However, the triplet loss and the quadruplet loss only consider the relative distances between samples, which is adequate for similarity measurements in tasks like person re-identification \cite{Quadruplet}, but is not helpful for HPV prediction task where the predicted score also needs to indicate the degree of HPV involvement. 

\subsection{Multiple Instance Learning pipeline}

As illustrated in Figure \ref{fig:Pipeline} \textbf{(A}), patches were firstly extracted from the tissue regions of the WSIs for training. The weakly labelling training strategy was used for training a deep neural network (i.e. the ResNet-18 \cite{ResNet} pre-trained on ImageNet was used in our experiments). The slide-level label (HPV status) was given to patches, and the cross entropy loss was used for loss calculation. After the base model was trained, the final fully connected layer was removed from the ResNet-18 model, and the rest of the network was used as a feature extractor for triplet ranking training. 

Bags were constructed based on slides and their corresponding labels, as shown in Figure \ref{fig:Pipeline} \textbf{(B)}. Let $B$ denote a set of patches extracted from the same slide. If the slide is HPV-, then we consider all the patches in $B$ can only present features of HPV- HNSCC. If the slide is HPV+, then a subset of the patches from $B$ can present the unique features of HPV+. Consider $\{B_1, B_2, ... , B_N\}$ to be $N$ bags containing patches extracted from $N$ slides, their corresponding labels $Y_n \in \{0, 1\}, n=1...N$ denote the slide-level label, where 0 represents HPV- and 1 represents HPV+. Within each bag $B$, there are K patches $\{I_{1}, I_{2}, ..., I_{K}\}$, where each patch is represented by a 512-dimensional feature vector $f \in \mathbb{R}^{512}$  extracted by the pre-trained ResNet-18 feature extractor. 

In each iteration of the triplet ranking training, two different HPV- bags and one HPV+ bag were randomly drawn from the dataset. The ResNet-18 feature extractor was used for generating feature embeddings of all the patches within a bag, and an MIL-classifier was used for generating a score for each patch within a bag. The MIL-classifier we used is a multilayer perceptron with one hidden layer, and one neuron on the output layer. A detailed structure of the MIL-classifier was illustrated in Figure \ref{fig:Pipeline} \textbf{(C)}. After generating scores for all the patches of the three bags, top K aggregation was used for aggregating a slide-level score from patch-level scores. In our design, the average of the top 10\% patches was used as the aggregation method. The proposed triplet ranking loss \eqref{eqn:OurLoss} was used for calculating the loss among three input bags.

\begin{table*}[!htbp]
	\begin{tabularx}{\linewidth}{
			>{\hsize=1.3\hsize}Y
			>{\hsize=.8\hsize}Y
			>{\hsize=.8\hsize}Y
			>{\hsize=.3\hsize}Y
			>{\hsize=.3\hsize}Y
			>{\hsize=.3\hsize}Y
			>{\hsize=.3\hsize}Y
		}
		\toprule
		& \multicolumn{2}{c}{3-fold CV on TCGA} 
		& \multicolumn{2}{c}{TCGA (n=412)} 
		& \multicolumn{2}{c}{Sheffield (n=69)} \\
		
		\cmidrule(lr){2-3} \cmidrule(l){4-5} \cmidrule(l){6-7}
		&AUC&AP&AUC&AP&AUC&AP \\
		\midrule
		Published result from \cite{KatherHPV} & 0.89 & - & - & - & - & - \\
		Klein et. al. \cite{KleinHPV} & 0.8829±0.0566 & 0.5992±0.1493 & 0.7961 & 0.4266 & 0.7905 & 0.8475 \\
		Kather et. al. \cite{KatherHPV} & 0.9035±0.0379 & 0.6106±0.1329 & 0.7936 & 0.3883 & 0.7733 & 0.8375 \\
		Lu et. al. \cite{CLAM} & 0.8600±0.0376 & 0.5882±0.0622 & 0.7176 & 0.3866 & 0.7647 & 0.7421 \\
		Campanella et. al. \cite{Campanella2019} & 0.596±0.0746 & 0.1913±0.0372 & 0.5975 & 0.1419 & 0.6111 & 0.6989 \\
		Ilse et. al. \cite{MaxWellingMIL} & 0.9019±0.023 & 0.6062±0.0885 & 0.5819 & 0.1314 & 0.8207 & 0.8726 \\
		\midrule
		TopK & 0.9035±0.0405 & 0.6189±0.1524 & 0.7864 & 0.4338 & 0.7922 & 0.8718 \\
		Cross Entropy & 0.8919±0.0468 & 0.5979±0.1117 & 0.8017 & 0.4659 & 0.7534 & 0.8053 \\
		MSE & 0.8617±0.0142 & 0.5771±0.0147 & 0.6855 & 0.1883 & 0.7621 & 0.8386 \\
		\textbf{Ours} & \textbf{0.9223±0.0397} & \textbf{0.7033±0.0938} & \textbf{0.8371} & \textbf{0.5463} & \textbf{0.8397} & \textbf{0.8737} \\
		\bottomrule
	\end{tabularx}
	\caption{Performance Comparison of different methods. The first two columns show the results of 3-fold cross validation experiments on TCGA cohort. The middle two columns show the results of using Sheffield cohort for training and TCGA cohort (n=412 patients) for testing.  The last two columns show the results of using TCGA cohort for training and Sheffield cohort (n=69 patients) for testing.}
	\label{Results}
\end{table*}

\section{Results}
\subsection{Experimental settings}
3-fold cross validation experiments were conducted on the TCGA-HNSC cohort for internal validation of our proposed method. Two external validation experiments were also conducted on TCGA-HNSC and Sheffield cohort, that is, use TCGA-HNSC as training and validation cohort, and report the test result on Sheffield cohort, and vice versa. For 3-fold cross validation, the dataset was split into 3 stratified folds. For external validation experiments, the training cohort was split into stratified 80\% for training, and 20\% for validation, and the result was reported on the testing cohort.

AUC (or AUROC, area under the receiver operating characteristic) was reported as the evaluation metric for our experiments. Considering the class imbalance issue of our dataset (e.g. TCGA-HNSC cohort only has 12\% HPV-positive rate), we also use the precision-recall curve (PRC) for comparison, and the average precision (AP) was reported for measuring the area under PRC (AUPR). For 3-fold cross validation experiments, the average and the standard deviation were reported. 

256 $\times$ 256 size patches were extracted under the 10$\times$ magnification level of WSIs. Stain and image augmentations were used when training the model, and stain normalisation \cite{Macenko} was used for inference.

\subsection{Comparison between different methods}
We have compared five published methods with the proposed method on HPV prediction problem. The first row present the baseline results reported in \cite{KatherHPV} with the same 3-fold cross validation setting. We re-conducted this baseline method with the ResNet-18 architecture, which achieved a better performance (Third row of Table \ref{Results}) than the published result. CLAM performed well on 3-fold CV, but did not generalise well on external validations. Surprisingly, \cite{Campanella2019} performed quite badly on all of the experiments. This might caused by the fact that the RNN aggregation used in their paper is hard to train with small amount of data and imbalanced dataset. The attention-MIL \cite{MaxWellingMIL} performed well on 3-fold CV and on Sheffield testing experiment, but performed badly on TCGA testing experiment. One possible reason can be that this method does not generalise well with only training on Sheffield (n=69) cohort.

To compare with other loss functions for this task, we have evaluated the performance of using cross entropy loss and MSE loss, as shown in Table \ref{Results}. To explored how much the triplet-MIL training is improving the results, we have used the patch scores predicted by weakly labelling ResNet-18 for aggregating the patient-level prediction, as shown in the 7th row of Table \ref{Results} (TopK method). A decrease in performance can be observed by removing the triplet-MIL training. This indicates that the triplet-MIL training process enables our MIL-classifier to learn essential features among all patches collectively, thus making better predictions on a patient-level.

\begin{figure*}[!ht]
	\begin{center}
		\includegraphics[width=1\textwidth]{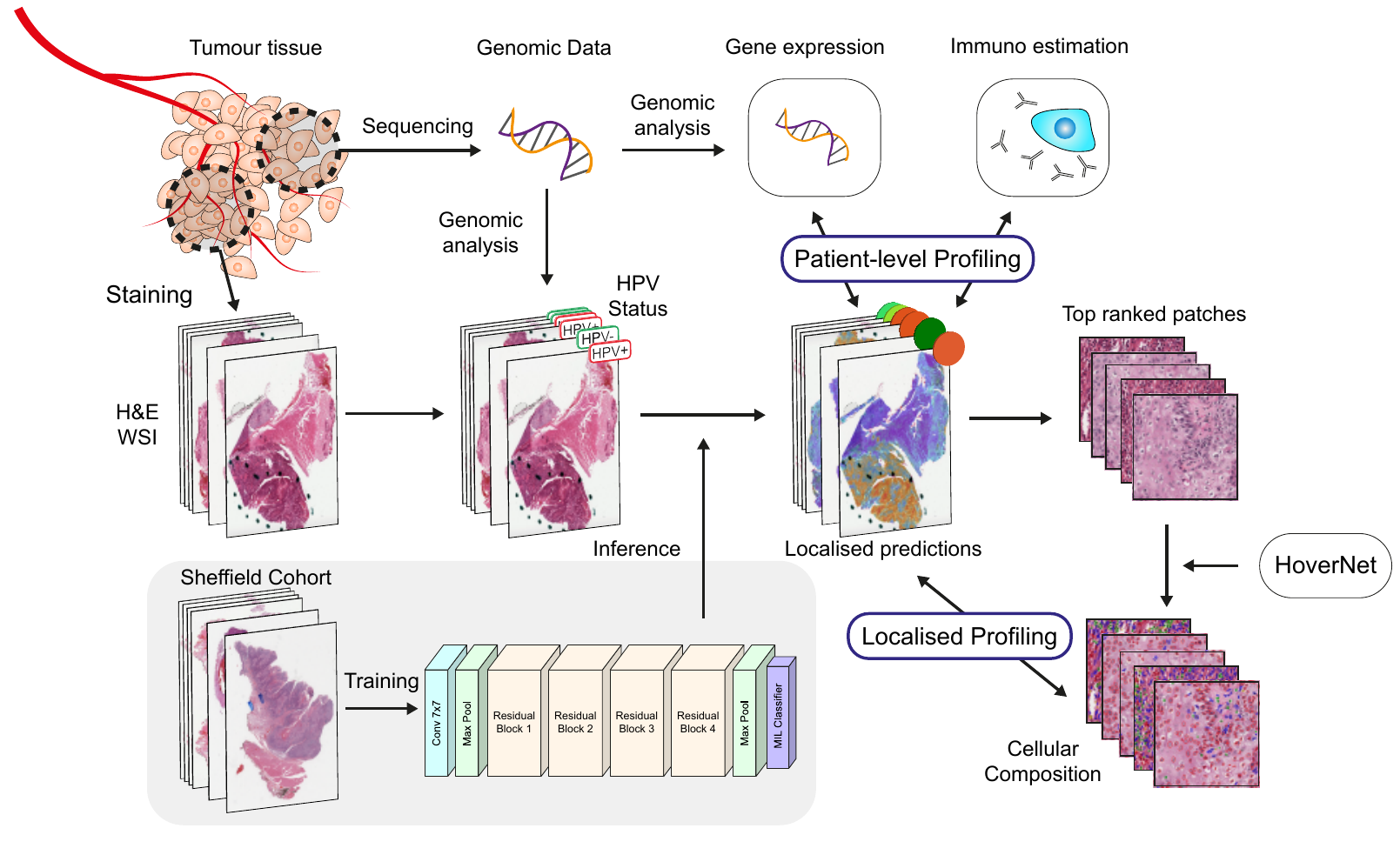}
	\end{center}
	\caption{Pipeline for our tumour microenvironment characterisation. The genomic and the H\&E WSI data was generated from different part of the tumour tissue. HPV status, gene expression levels and the immune cells estimations were all generated from the gene sequencing data. A proposed model trained on the Sheffield cohort was used on TCGA cohort for generating scores. A patient-level profiling was conducted using patient-level score and patient-level genomic and immunological data. A localised profiling was conducted using localised prediction scores and localised cellular composition data.}
	\label{fig:Inference}
\end{figure*}

\begin{figure*}[!ht]
	\begin{center}
		\includegraphics[width=1\textwidth]{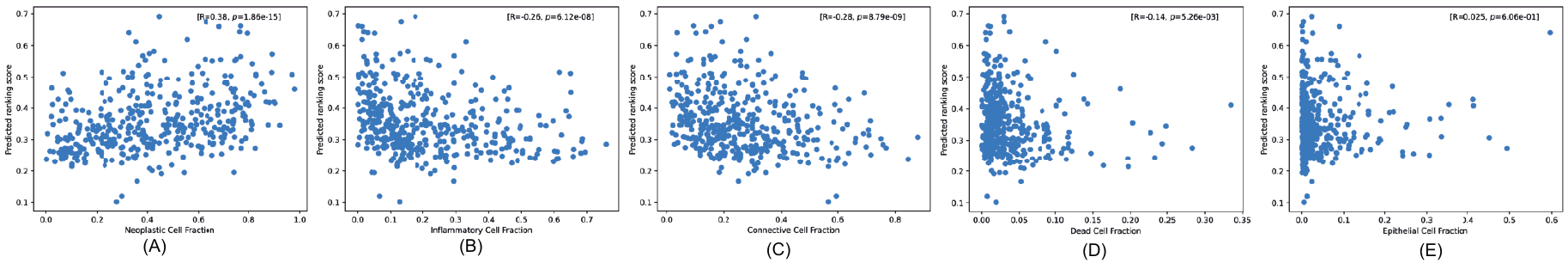}
	\end{center}
	\caption{Pearson correlation between the predicted ranking score and detected cell fractions. (A) Neoplastic Cell Fraction, (B) Inflammatory Cell Fraction, (C) Connective Cell Fraction, (D) Dead Cell Fraction, (E) Epithelial Cell Fraction.}
	\label{fig:CellCompCorr}
\end{figure*}

\begin{figure*}[!ht]
	\begin{center}
		\includegraphics[width=1\textwidth]{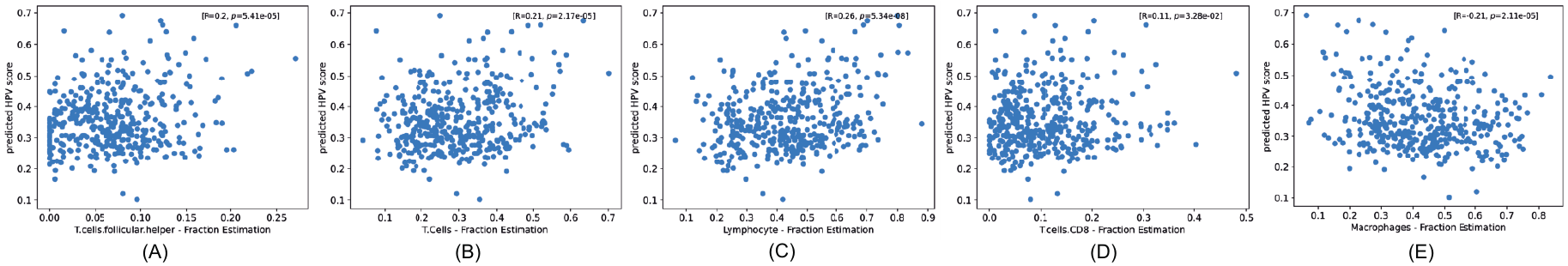}
	\end{center}
	\caption{Pearson correlation between the predicted ranking score and estimated immune cell fractions. (A) TFH Fraction Estimation, (B) T Cell Fraction Estimation (C) Lymphocyte Fraction Estimation, (D) CD8+ T Cell Fraction Estimation, (E) Macrophages Fraction Estimation.}
	\label{fig:ImmuneCorr}
\end{figure*}

\subsection{Characterisation of the tumour microenvironment between HPV+/- HNSCC patients with our ranking score}
In this study, we conducted a comprehensive characterisation analysis of the tumour microenvironment in the TCGA cohort. A ranking model trained on Sheffield cohort 
was used for inference on the whole TCGA cohort, generating an ranking score for each patient. We performed both a patient-level profiling and a localised profiling, as illustrated in Figure \ref{fig:Inference}. Our patient-level profiling characterised the HPV's impact on the tumour microenvironment from a genetic and immunological level using the predicted ranking score. Since the genomic data from TCGA was sequenced from the tissue block which is different from the one used to generate the diagnostic slide, some differences may exist in the tumour microenvironment between these two specimens. Therefore, this information is regarded as patient-level profile. Although we do not have the localised genomic data for our diagnostic slides, a localised cellular composition analysis was performed among highly ranked patches. This can give us insights into the relations between the HPV infection and the tumour microenvironment of the WSI data we used to make predictions.

\begin{table}[!h]
	\begin{center}
		\begin{tabular}{c|c c}
			\hline
			Cell types & rho & \textit{p} value \\
			\hline
			Neoplastic& 0.38 & 1.86 $\times$ $10^{-15}$  \\
			Inflammatory& -0.26 & 6.12 $\times$ $10^{-8}$  \\
			Connective& -0.28& 8.79 $\times$ $10^{-9}$  \\
			Dead& -0.14 & 5.2 $\times$ $10^{-3}$  \\
			Epithelial& 0.03 & 0.61 \\
			\hline
		\end{tabular}
	\end{center}
	\caption{Pearson correlation between our ranking score and fractions of identified cell types on top ranked patches.}
	\label{CellComposition}
\end{table}
\subsubsection{Cellular composition profiling of HPV+/- HNSCC}
To understand the impact of HPV infection on the tumour microenvironment in a cellular-level, as well as to interpret what the model learnt during training, a localised cellular composition analysis was conducted on the TCGA cohort. The nuclei detection model HoverNet \cite{Hovernet} trained on the PanNuke dataset \cite{PanNuke} was used to identify 5 types of cells (i.e. Neoplastic, Epithelial, Inflammatory, Connective, Dead) on patches with high ranking scores. Fractions of each cell type was calculated on top patches, and Pearson correlation analysis was performed. The detailed subtypes of these 5 categories of cells can be checked in the work of \cite{PanNuke}. 


As shown in Table \ref{CellComposition}, a positive correlation with neoplastic cell fraction was identified, and negative correlations were identified with inflammatory, connective and dead cell fractions. The correlation with the epithelial cell fraction is not significant. A high correlation between our score and the neoplastic cell fraction may indicates that the distinct features for HPV determination lie in patches with more tumour cells. 

\begin{figure*}[!ht]
	\begin{center}
		\includegraphics[width=1\textwidth]{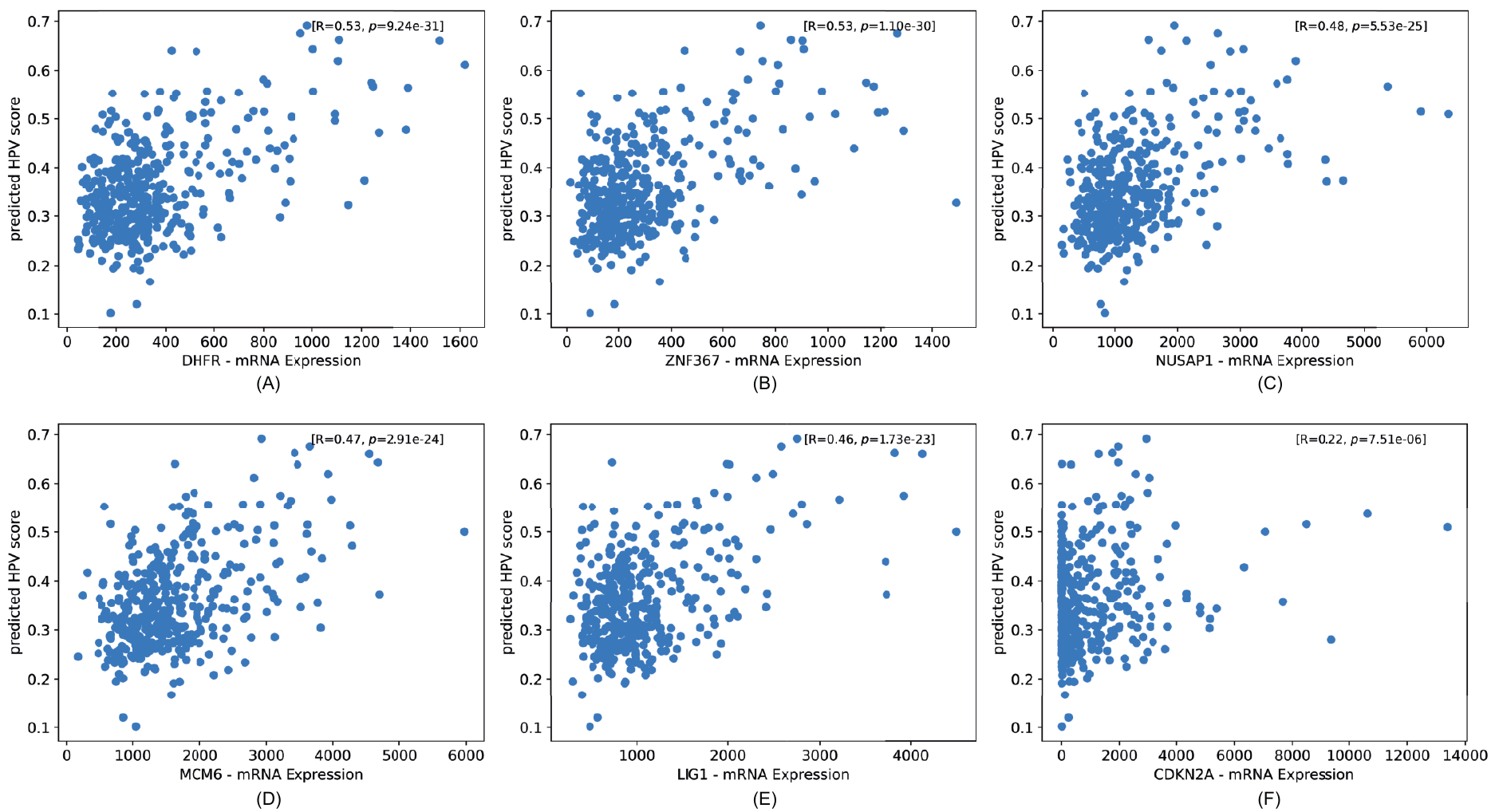}
	\end{center}
	\caption{Pearson correlation between the predicted ranking score and gene expression levels. (A) DHFR mRNA Expression, (B) ZNF367 expression (C) NUSAP1 expression, (D) MCM6 expression, (E) LIG1 expression, (F) CDKN2A expression}
	\label{fig:mRNA}
\end{figure*}

Meanwhile, negative correlations of inflammatory cells (including lymphoid and macrophage cells in PanNuke categories) and connective cells (including fibroblasts, endothelial, myofibroblasts cells,  etc. in PanNuke categories) with the predicted ranking scores have been identified. This means the higher our ranking score is, the less inflammatory cell and connective cell will be present in the top ranked patches. From a clinical perspective, an increased expression level of tumour-associated macrophages \cite{macrophages_prognosis} and an amplified tumour-associated fibroblast presence \cite{fibroblast_prognosis} have been found to be correlated with a poor prognosis in HNSCC, respectively. This might indicate the HPV infection can have an impact on the involvement of macrophages and fibroblasts in the tumour microenvironment. Moreover, causing prognostic differences of HPV+/- HNSCCs.

\begin{table}[!h]
	\begin{center}
		\begin{tabular}{c|c c}
			\hline
			Immune Cell types & rho & \textit{p} value \\
			\hline
			TFH & 0.28 & 3.96 $\times$ $10^{-9}$  \\
			T cells & 0.27 & 1.98 $\times$ $10^{-8}$  \\
			Lymphocytes & 0.26& 4.94 $\times$ $10^{-8}$  \\
			CD8+ T cells & 0.24 & 1.04 $\times$ $10^{-6}$  \\
			Macrophages& -0.20& 6.0 $\times$ $10^{-5}$ \\
			\hline
		\end{tabular}
	\end{center}
	\caption{Pearson correlation between our ranking score and the estimated fractions of immune cell types.}
	\label{ImmuneFrac}
\end{table}

\subsubsection{Immunology profiling of HPV+/- HNSCC}
Furthermore, an immunology profiling of HPV+/- HNSCC patients using the proposed ranking score was conducted in this study. Correlation analysis was performed between the estimated fraction of immune cells and the corresponding ranking score. As shown in Table \ref{ImmuneFrac}, positive correlations were identified between our ranking score with the fractions of the T cells follicular helper (TFH), T cells, lymphocytes, and CD8+ T cells. A negative correlation was identified with the macrophages fraction. Based on clinical findings, a higher expression level of TFH is associated with a favourable prognosis in HNSCC patients \cite{cillo_immune_2020}. A negative correlation with the macrophages fraction is also inline with our cellular composition analysis, where a negative correlation with the fraction of inflammatory cells was identified. These TME characteristics might be the underlying reason or the causation to the better prognosis of HPV+ HNSCC. 

\begin{table}[!h]
	\begin{center}
		\begin{tabular}{c|c c c}
			\hline
			Gene ID & rho & \textit{p} value & Behaviour \\
			\hline
			DHFR & 0.53 & 9.24 $\times$ $10^{-31}$  & Upregulated\\
			ZNF367 & 0.53 & 1.1 $\times$ $10^{-30}$ & Upregulated \\
			NUSAP1 & 0.48& 5.53 $\times$ $10^{-25}$ & Upregulated \\
			MCM6 & 0.47 & 2.91 $\times$ $10^{-24}$  & Upregulated \\
			LIG1& 0.46& 1.73 $\times$ $10^{-23}$ & Upregulated \\
			...& ...& ... & ... \\
			CDKN2A& 0.22& 7.51 $\times$ $10^{-6}$ & Upregulated \\
			\hline
		\end{tabular}
	\end{center}
	\caption{Pearson correlation between our ranking score and the mRNA expression level of top 5 significantly correlated genes and CDKN2A. Behaviour column indicates whether this gene has been upregulated or downregulated in HPV+ samples comparing to HPV- samples, which was reported in the work of \cite{gene_expression}}
	\label{GeneExpression}
\end{table}

\subsubsection{Genomic profiling of HPV+/- HNSCC}
To investigate the HPV's impact on the genomic landscape of the tumour microenvironment between HPV+/- HNSCC, as well as to investigate whether this difference in gene expressions, if any, can be reflected by our ranking score, a genomic profiling was performed on TCGA cohort. Pearson correlation analysis was used to find the relations between the mRNA expression data with the predicted ranking scores. To investigate whether the correlations is in line with the clinical findings, we compared the genes we identified with the differentially expressed genes published in \cite{gene_expression}. 123 genes have been identified as positively correlated ($p<0.05$) with our ranking scores, and were reported as being upregulated in \cite{gene_expression}, and 7 genes have been identified as negatively correlated ($p<0.05$), and were reported as being downregulated in \cite{gene_expression}. Top 5 of these genes can be seen in Table \ref{GeneExpression}. Some of these genes (e.g. NUSAP1, MCM6, LIG1) are responsible for regulating the cell lifecycle, nuclei acid metabolism, DNA replication, repair, and recombination (DRRR), and Cellular assembly and organization \cite{gene_expression}. This might indicate the HPV-induced carcinogenesis has more effect on these cell-lifecycle regulating genes due to the distinct mechanism of HPV infection, whereas non-viral driven carcinogenesis has less impact on the disregulation of these genes.

A moderate but significant positive correlation was also identified with the gene CDKN2A which encodes p16, as shown in Figure \ref{fig:HPVMechanism}. This is in line with the clinical findings that HPV infection can cause an overexpression of p16. The reason why this correlation is not strong might be that the H\&E slides we used for training our model do not contain abundant information for determining the p16 status. Nevertheless, the correlations we identified matches the clinical observations, and validates the effectiveness of our proposed model.

\section{Discussion}
In this paper, we proposed a novel ranking loss function and an MIL pipeline for HPV status prediction in HNSCC patients using the H\&E-stained WSIs. Internal and external validations were conducted on multicentric TCGA-HNSC cohort and single-centre Sheffield cohort. Comparisons with simple weakly labelling methods, other MIL-based methods, other loss functions and the ablation study have shown the effectiveness of the proposed method. 

Furthermore, a characterisation of HPV's impact on the tumour microenvironment has been performed. We explored three aspects of the TME: genomic profile, immunological profile and cellular composition, and from both patient-level and local-level. The genomic profiling has discovered some significant correlations between our ranking score and gene expression level, which is in line with the clinical findings. From our immunological profiling, a possible link between HPV infection and the involvement of TFH and the CD8+ T cells, and their prognostic significance.

The genomic and the immunological profiling have revealed the possible patterns of the HNSCC tumour microenvironment caused by HPV involvement in a molecular level. However, since the WSI data we used for inference was not the exact image of the tissue which was used for generating the genomic data, there might be discrepancies in the tumour microenvironment between the sequencing specimen and the scanned H\&E slides. Therefore, we conducted a localised profiling of the cellular composition of top ranked patches. We found a positive correlation with the neoplastic cell fraction, and negative correlations with inflammatory, connective and dead cells, which is also in accordance with clinical findings. This indicates a possible impact of HPV infection on the abundance of these cells, which can lead to different survival outcome. Meanwhile, it reveals what features our model might have taken during the learning process.

A better survival outcome of HPV+ HNSCC has revealed the importance of patient stratification based on HPV infection status. Meanwhile, more studies have shown the prognostic value of immunology profiles \cite{spanos_immune_2009, wansom_correlation_CD8_2010, macrophages_prognosis, fibroblast_prognosis, cillo_immune_2020} and genomic patterns \cite{gene_expression}. This can enable us to discover more potential biomarkers for HNSCC patient management. In this study, only the gene expression data was analysed. However, more evidence on the role of miRNA interaction \cite{role_mirna, zheng_regulation_2011} and DNA methylation \cite{worsham_epigenetic_2013, von_knebel_doeberitz_role_2019} in HPV-induced HNSCC have been reported, and undoubtedly worth further investigation. This can lead us to a better understanding of the HPV-induced carcinogenesis, as well as developing better treatment for HNSCC patients in clinical practice.

\section*{Acknowledgments}
This study was partly supported by a PhD studentship to the first author funded by the General Charities of the City of Coventry. The first author is also grateful to the Computer Science Doctoral Training Centre at the University of Warwick for providing funding for this research. The results published here are in part based upon data generated by the TCGA Research Network: https://www.cancer.gov/tcga.

\bibliographystyle{plain}  
\bibliography{main}  

\end{document}